\documentclass[conference]{IEEEtran}
\usepackage[OT1]{fontenc} 
\usepackage[noadjust]{cite}
\usepackage{siunitx}
\makeatletter
\def\input@path{{fig/}} 
\makeatother
\usepackage[pdftex]{graphicx}
\usepackage{amsmath}
\usepackage{booktabs}
\usepackage{array}
\usepackage{xcolor}
\usepackage{listings}
\usepackage{comment}
\usepackage[usestackEOL]{stackengine}
\usepackage[caption=false,font=footnotesize]{subfig}
\usepackage{url}
\usepackage{amssymb}
\usepackage{threeparttable}

\newcommand{\subparagraph}{}
\usepackage{titlesec}
\usepackage{arydshln}
\usepackage{pifont}
\titleformat{\paragraph}[runin]{\normalfont\bfseries}{}{0pt}{}[. \mbox{}]

\def\BibTeX{{\rm B\kern-.05em{\sc i\kern-.025em b}\kern-.08em
    T\kern-.1667em\lower.7ex\hbox{E}\kern-.125emX}}

\newcommand{\figrtext}{Fig.}
\newcommand{\tabrtext}{Table}

\newcommand{\secrtext}{Section}
\newcommand{\figref}[1]{\figrtext~\ref{fig:#1}}

\newcommand{\tabref}[1]{\tabrtext~\ref{tab:#1}}

\newcommand{\secref}[1]{\secrtext~\ref{sec:#1}}

\newcommand{\etal}{\textit{et al}. }
\newcommand{\ie}{\textit{i}.\textit{e}., }
\newcommand{\eg}{\textit{e}.\textit{g}., }

\newcommand{\stimes}{\mathsf{x}}
\begin{document}

\title{GrateTile: Efficient Sparse Tensor Tiling for\\CNN Processing\vspace{-.5em}}

\author{
\IEEEauthorblockN{Yu-Sheng Lin}\IEEEauthorblockA{\textit{\qquad Inventec Corporation\qquad}\\ lin.john@inventec.com}
\and
\IEEEauthorblockN{Hung-Chang Lu}\IEEEauthorblockA{\textit{\qquad National Taiwan University\qquad}\\ hclu@media.ee.ntu.edu.tw}
\and
\IEEEauthorblockN{Yang-Bin Tsao}\IEEEauthorblockA{\textit{\qquad National Taiwan University\qquad}\\ yangbin@media.ee.ntu.edu.tw}
\and
\IEEEauthorblockN{Yi-Min Chih}\IEEEauthorblockA{\textit{\qquad National Taiwan University\qquad}\\ yiminchi@media.ee.ntu.edu.tw}
\and
\IEEEauthorblockN{Wei-Chao Chen}\IEEEauthorblockA{\textit{\qquad Inventec Corporation\qquad}\\ chen.wei-chao@inventec.com}
\and
\IEEEauthorblockN{Shao-Yi Chien}\IEEEauthorblockA{\textit{\qquad National Taiwan University\qquad}\\ sychien@media.ee.ntu.edu.tw}
}
\maketitle

\begin{abstract}
We propose GrateTile, an efficient, hardware-friendly data storage scheme for sparse CNN feature maps (activations).  It divides data into uneven-sized subtensors and, with small indexing overhead, stores them in a compressed yet randomly accessible format.  This design enables modern CNN accelerators to fetch and decompressed sub-tensors on-the-fly in a tiled processing manner.  GrateTile is suitable for architectures that favor aligned, coalesced data access, and only requires minimal changes to the overall architectural design.  We simulate GrateTile with state-of-the-art CNNs and show an average of 55\% DRAM bandwidth reduction while using only 0.6\% of feature map size for indexing storage.
\end{abstract}

\IEEEpeerreviewmaketitle

\section{Introduction}
Convolutional neural networks~(CNNs) are now considered one of the most widely used machine learning techniques in computer vision and image processing~\cite{vdsr,alexnet,resnet,vgg,googlenet}.  Its primary operation is the convolution between kernels~(weights) and feature maps~(activations), which can consume lots of power through MAC operations and memory accesses.  To alleviate this problem, one can take advantage of the redundancies in the feature maps and skip unnecessary processing with \emph{sparse} computation.  For example, neural networks using the ReLU activation function may have highly sparse feature maps with up to 80\% zero values clipped from negative values.  It is also possible to fine-tune the network to generate kernels with higher sparsity~\cite{eie, Liu_2015_CVPR}, so that CNN accelerators can reduce operation waste by gating the processing elements~(PEs) and avoid scheduling zero operations~\cite{eyeriss,eie,scnn}.

\begin{figure}[t]%
\centering
\includegraphics[width=0.42\textwidth,trim={0 4em 0 0}]{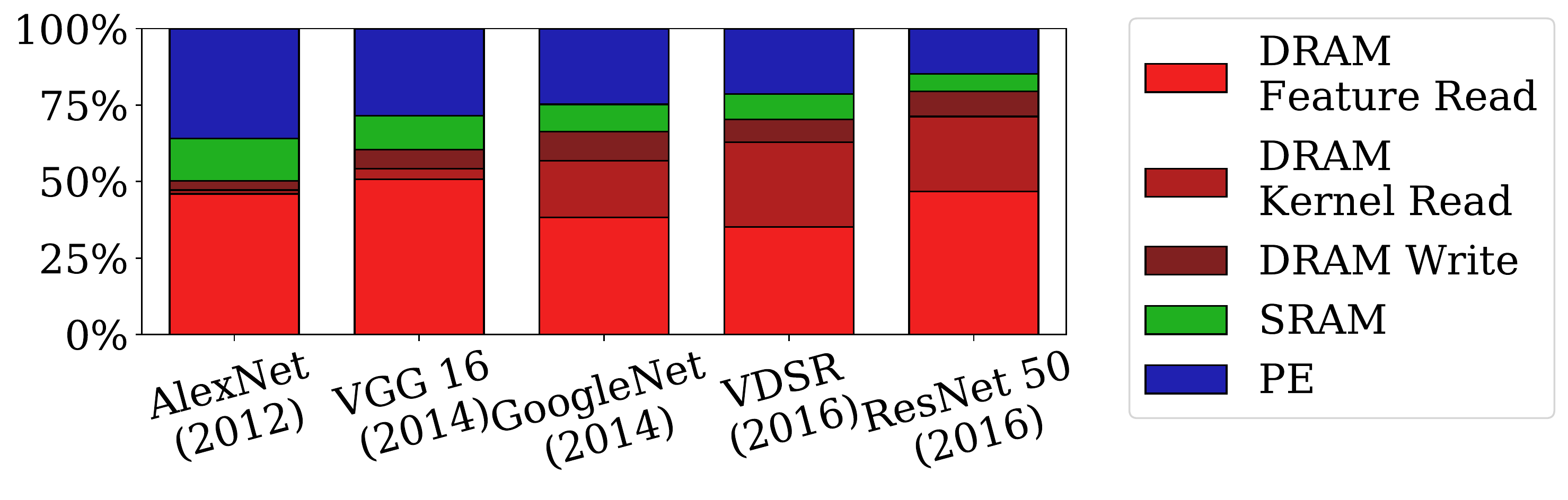}%
\caption{\textbf{Power breakdown of popular CNN applications using SCALE-sim.}  According to this simulation, the DRAM feature read is the primary power draw for CNNs.}\label{fig:scalesim}
\end{figure}

\begin{figure}[t]%
\centering
\subfloat[Tiled CNN processing without compression is inefficient.]{\includegraphics[scale=0.41,page=1]{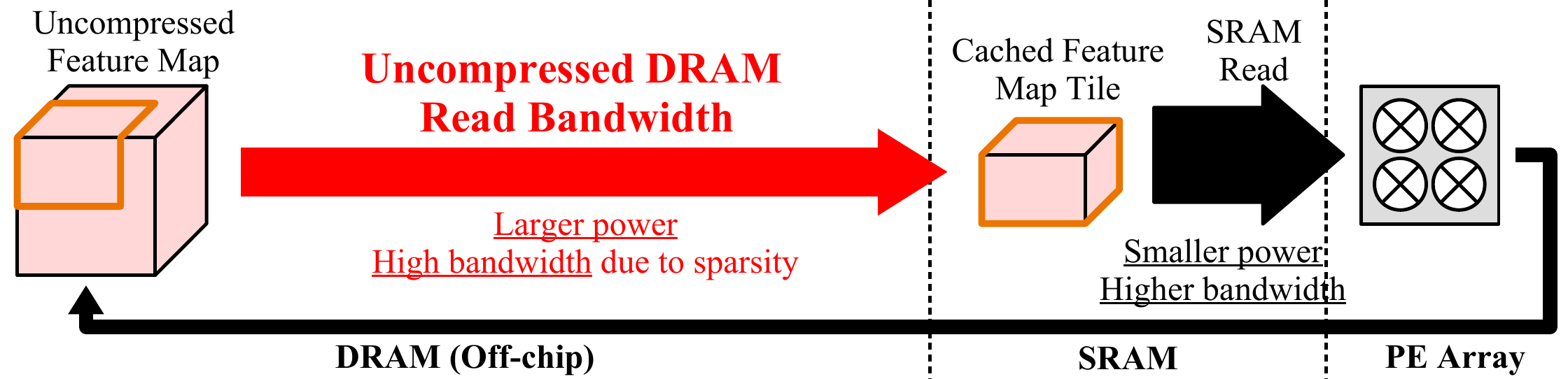}\label{fig:cover0:a}}\\
\subfloat[Compressed feature maps do not work well with tiling.]{\includegraphics[scale=0.41,page=2]{cover/cnn_overview.pdf}\label{fig:cover0:b}}\\
\subfloat[Independently compressed subtensors are tiling-friendly.]{\includegraphics[scale=0.41,page=3]{cover/cnn_overview.pdf}\label{fig:cover0:c}}%
\caption{\textbf{Reducing DRAM bandwidth via compression.} While it can be more power-efficient to compressed feature maps in DRAMs, traditional compression algorithms tend to be incompatible with tiled processing.  Dividing feature maps into subtensors and compress them independently is an effective method for this purpose.}\label{fig:cover0}
\end{figure}

Compared with the energy wasted through redundant operations, data access power is arguably more critical for future accelerator designs because memory bandwidth has been growing slower than the speed of PEs~\cite{memwall}.  That is, an algorithm can become increasingly memory bound for future architectures.  Newer networks tend to adopt smaller convolution kernels with deeper layers, which further reduces operation count at the cost of increased memory usage.  In \figref{scalesim}, we calculate the power consumption breakdown according to \cite{dram_power} by simulating several popular CNNs with SCALE-sim on a $16\times 16$ systolic array~\cite{scalesim,tpu,systolic}.  Notice that the percentage of MAC power decreases from $35$\% in 2012 to $15$\% in 2016, while the DRAM feature read consistently consumes over half of the remaining power.  Modern CNN accelerators have already utilized on-chip SRAM to effectively reduce data access power~(\figref{cover0:a}).  To push the envelope further, we could compress the feature maps, but the compression scheme may not be compatible with the tiled processing nature of modern CNN accelerators~(\figref{cover0:b}). A better approach is to divide the feature maps into independently compressed \emph{subtensors} to make them randomly-accessible for tiled processing~(\figref{cover0:c}).  This design principle allows the memory controller to fetch only the required subtensors and assemble them into tiles on-the-fly, without wasting bandwidth on over-fetching data outside of the tiles.

\begin{figure}[t]%
\centering
\subfloat[Divide feature map into large subtensors.]{\label{fig:cover1:a}\includegraphics[trim={0 1cm 0 0},scale=0.41]{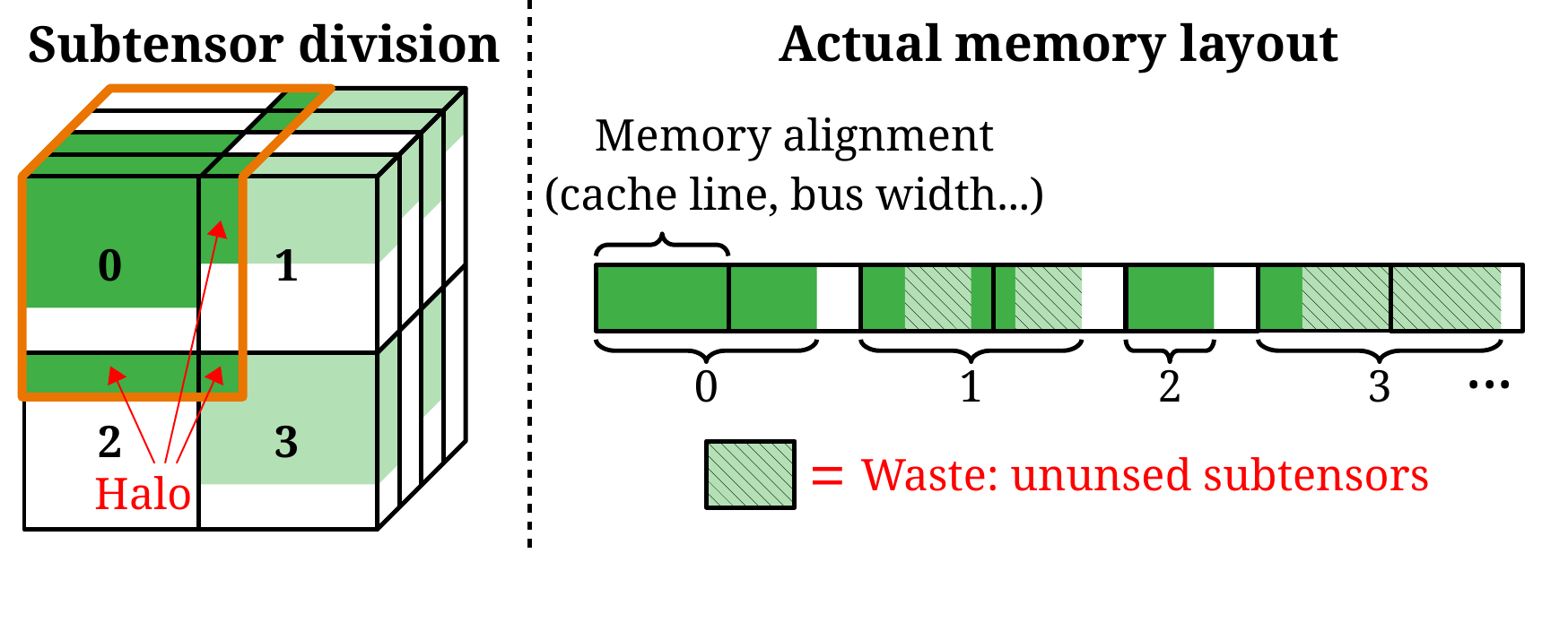}}\\
\subfloat[Divide feature map into small subtensors.]{\label{fig:cover1:b}\includegraphics[scale=0.41,page=2]{cover/overfetch.pdf}}\\
\subfloat[GrateTile feature map division.]{\label{fig:cover1:c}\includegraphics[trim={0 1cm 0 0},scale=0.41,page=3]{cover/overfetch.pdf}}
\caption{\textbf{A comparison of feature map division methodologies.}  The use of large subtensor sizes~\cite{armmlp} can cause partial subtensor accesses. On the other hand, small subtensor sizes~\cite{cambriconx} can cause data fragmentation. Both cases lead to bandwidth wastes. GrateTile uses hybrid subtensor sizes aligned to the CNN data fetching pattern, preventing both types of problems at the same time.}\label{fig:cover1}
\end{figure}

After analyzing how CNN architectures divide, compress, and store the feature maps~\cite{armmlp,cambriconx}, we observe that the division and storage process has an equally if not more substantial impact on the overall DRAM bandwidth when compared with the compression algorithms for individual subtensors.  \figref{cover1} illustrates the trade-off between using a larger subtensor size, which causes wasted fetch (\figref{cover1:a}), and a smaller subtensor size which causes data fragmentation (\figref{cover1:b}).  To break free from this trade-off, we propose GrateTile, which divides the feature maps into uneven sizes for optimal CNN processing (\figref{cover1:c}).  By inserting smaller subtensors between larger tensors, GrateTile combines the storage efficiency of larger subtensors without the over-fetching waste.  By adding GrateTile functionality to existing CNN accelerators, we can gain approximately $55$\% bandwidth improvement over the uncompressed baseline, and $6$-$27$\% bandwidth improvement over compressed tiles according to our simulation.  In summary, our contributions are:
\begin{itemize}
\item A bandwidth-efficient storage scheme for sparse feature maps with CNN accelerator-friendly memory access patterns,
\item A universal methodology to convert a sparse tensor into the GrateTile packing given CNN layer and accelerator configurations, and
\item A method for integrating GrateTile into existing accelerators with small hardware modification and overhead.
\end{itemize}

\section{Related Works}\label{sec:relate}

While GrateTile focuses on exploiting feature map sparsity that tends to be dynamic, kernel sparsity tends to be more static.  Researchers have gained great success exploiting this attribute to reduce DRAM bandwidth and power consumption for CNNs.  These methods operate by dropping small kernel values followed by retraining to compensate for accuracy loss.  Many of these works also focus on designing PEs that can skip unnecessary MAC or math operations to save power.  EIE~\cite{eie} is a fully-connected layer accelerator, which uses two indices for the next non-zero values in the feature maps and kernels so that it only performs operations with non-zero operands.  Several CNN accelerators also apply similar methods to skip unnecessary operations~\cite{eyeriss,cnvlutin}.  This processing flow forces serialization of operation for different kernel values and therefore limits the parallelism.  SCNN decomposes a matrix multiplication or a CNN into several outer product operations~\cite{scnn} and is thus an outer product PE array.  Since the outer product is zero when either kernel or feature map is zero, it compacts the input to ensure there is no wasted operation.  However, since the output address from each PE is different, this causes irregular address calculation and results in a distribution unit that is three times larger than the PE array.

The methods above assume the zero values can appear randomly.  On the other hand, some researchers believe the sparsity can be \textit{structural}, and subtensors of kernel tensors may repeatedly appear in different positions.  CirCNN and PermDNN assume any row in the kernel tensor is a rotation of its neighboring row~\cite{permdnn,circnn0}.  These repeating structure of kernel tensors enable many methods for reducing operation count, such as replacing the multiplication by table lookup.  Wu~\etal use the \textit{vector quantization} to cluster kernels by $k$-means and replace them by their cluster center indices to save memory~\cite{vq}.  

\section{GrateTile for Sparse Feature Maps}\label{sec:propose}

\begin{figure}[t]%
\centering
\subfloat[Bitmask compression.]{\label{fig:compression:a}\includegraphics[scale=0.4,page=2]{cover/cnn_compress.pdf}}\qquad%
\subfloat[ZRLC compression.]{\label{fig:compression:b}\includegraphics[trim={0 0 1.75cm 0},scale=0.4,page=1]{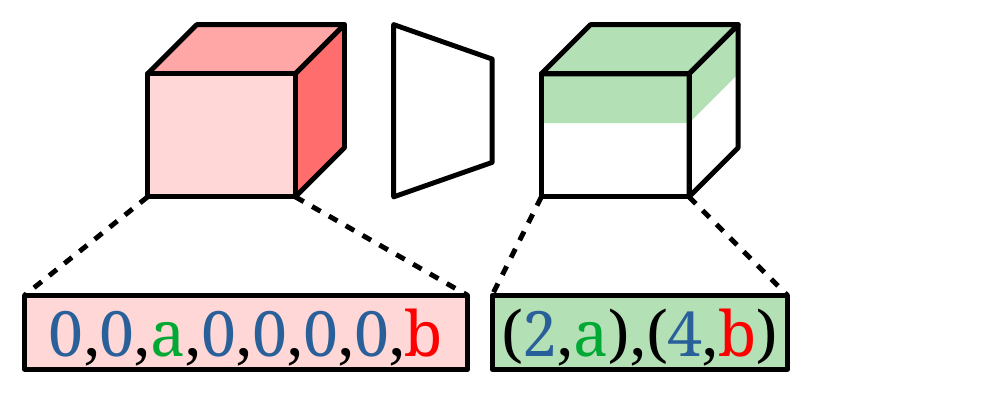}}
\caption{\textbf{Two popular methods for CNN feature map compression.}  Bitmask and ZRLC are widely used algorithms in CNN accelerators due to their simplicity for hardware implementation.}\label{fig:compression}
\end{figure}

\begin{figure*}[t]
\centering
\subfloat[Processing the first $8\times 8$ CNN block.]{\label{fig:gratetile:a}\includegraphics[scale=0.4,page=1]{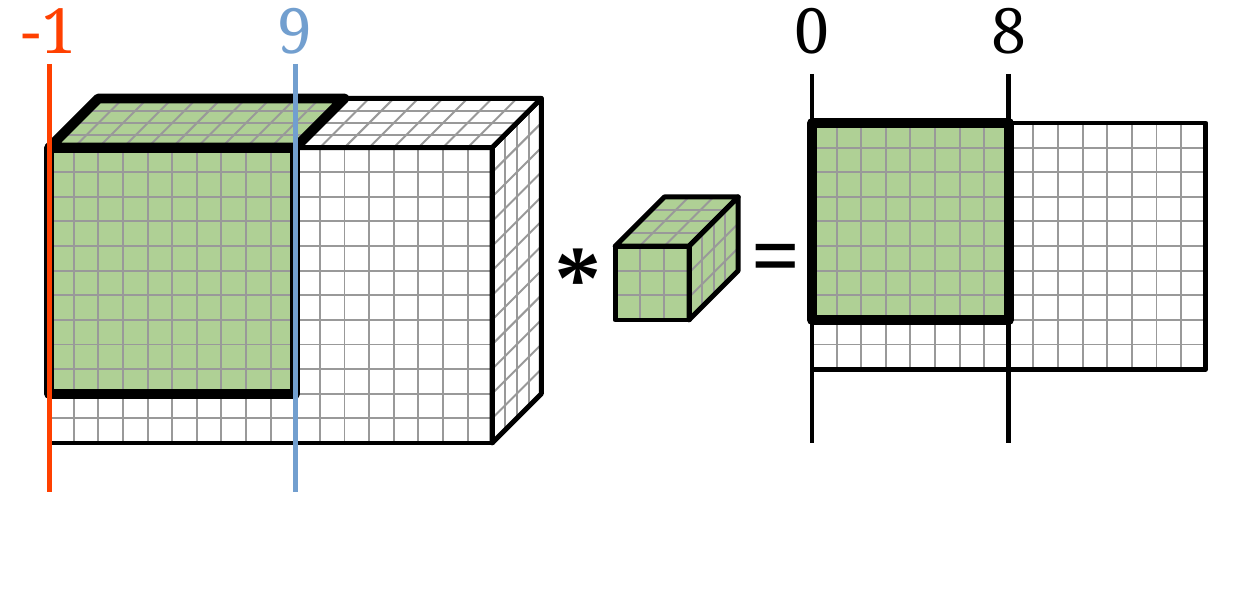}}%
\subfloat[Processing the second $8\times 8$ CNN block.]{\label{fig:gratetile:b}\includegraphics[scale=0.4,page=2]{propose/gratetile.pdf}}\quad%
\subfloat[GrateTile.]{\label{fig:gratetile:c}\includegraphics[scale=0.46,page=3]{propose/gratetile.pdf}}%
\caption{\textbf{Deriving the GrateTile Configuration.}  The proposed methodology divides the feature map tensor along all possible boundaries of the accessed tensors using modular arithmetic.}\label{fig:gratetile}
\end{figure*}

\subsection{The Need for Hardware Aligned Storage}
As discussed before in \figref{cover1}, to support tiled CNN processing,
we need to divide the feature maps into subtensors and compress them independently.  Many architectures adopt simple compression algorithms such as bitmask or zero run-length coding~(ZRLC)~\cite{armmlp,cnvlutin,eyeriss,cambriconx} with uniform division scheme for the tensors~(\figref{compression}).  While suitable for hardware implementation, this can lead to a waste of memory bandwidth.  Even though the convolution operation only requires data from surrounding pixels~(\ie halo), we may end up fetching the entire neighboring subtensors because the compressed blocks are not randomly accessible~(\figref{cover1:a}).  Furthermore, modern memory hierarchies, like DRAM or cache, favor aligned and coalesced access, and the variable size of compressed data can result in fragmentation and wasted bandwidth.  We can reduce fragmentation with index memory~(\figref{cover1:b}), but this pointer index can be too big for the on-chip SRAM, or contribute to additional latency and bandwidth if stored in the DRAM

\figref{cover1:c} shows how GrateTile can eliminate both the aforementioned problems.  By unevenly dividing the subtensors, we have fewer subtensors and smaller index memory sizes, while ensuring proper boundary alignment for tiled processing.  Next, we explain the methodology for finding an optimal division given a CNN and the hardware configuration.

\begin{figure}[t]%
\centering
\subfloat[]{\includegraphics[page=1,scale=0.3]{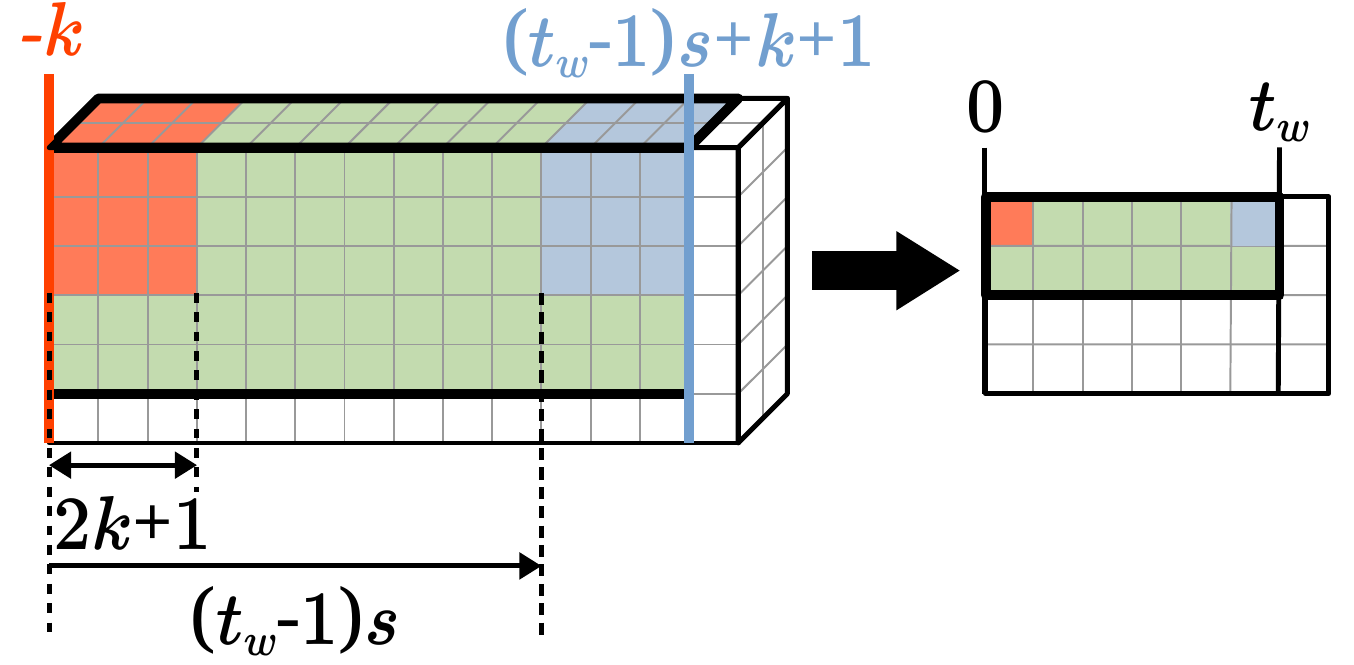}\label{fig:gratetile_general:a}}\quad
\subfloat[]{\includegraphics[page=2,scale=0.3]{propose/gratetile_general.pdf}\label{fig:gratetile_general:b}}%
\caption{{\bf Deriving GrateTile configuration for CNN layers.} (a) A standard CNN with $(k,s,d,t_w)=(1,2,1,6)$ and (b) a dilated CNN with $(k,s,d,t_w)=(1,1,2,6)$.}\label{fig:gratetile_general}%
\end{figure}

\subsection{Computing GrateTile Configuration}
Consider a CNN architecture processing a $3\times 3$ convolution on $4$ input channels, using an $8\times 8$ tile size for the output feature map~(\figref{gratetile}).  Our goal is to create a feature map division that (1) avoids accessing partially compressed subtensors, and (2) minimizes the number of subtensors to reduce data fragmentation.  In this example, to compute the first output tile, we need to fetch a $10\times 10\times 4$ input tile~(\figref{gratetile:a}).  When processing the next output tile, we would step toward the right by $8$ elements on the feature map~(\figref{gratetile:b}) to fetch the next input tile.  Since the step size is constant within one layer of CNN processing, the left~(orange) and right~(cyan) access boundaries form two arithmetic progressions, denoted as $\mathcal{B}_l = \left\lbrace-1,7,15\cdots\right\rbrace$ and  $\mathcal{B}_r=\left\lbrace9,17,25\cdots\right\rbrace$.  The GrateTile configuration is simply the divisions formed by both bounaries, namely the union $\mathcal{G} = \mathcal{B}_r \cup \mathcal{B}_l$, or simply $\mathcal{G} = \left\lbrace 1, 7 \right\rbrace \pmod 8$, as shown in \figref{gratetile:c}.  Because $7-1 = 6$ and $1-7 = 2 \pmod 8$, each spatial dimension of the feature map is divided into two uneven sizes of $2$ and $6$, which results in four subtensor shapes---$6\times 6$, $2\times 6$, $6\times 2$, and $2\times 2$.  A $10\times 10$ window is then composed of one $6\times 6$, two $2\times 6$ and $6\times 2$, and four $2\times 2$ subtensors.  Also, since the halo only appears in the spatial dimension, this division process is not necessary along the channel dimension.

We now generalize this example for all modern CNN layers, whose computations can be defined the following three parameters:

\begin{itemize}
\item Kernel size---denoted as $2k+1$ since kernel sizes tend to be odd integers.
\item \emph{Two output elements} convolving \emph{two windows} with a stride of $s$.  When $s>1$, it means a smaller output feature map and thus less computation cost.
\item Dilated CNN~\cite{dilconv} convolves strided input elements for \emph{one output element} to enlarge the equivalent window size, and we denote this stride as $d$.
\end{itemize}

Besides, we denote the output tile size as $t_h\times t_w$.  \figref{gratetile_general:a} shows a CNN with $d=1$.  To compute the leftmost output element, we fetch from the feature map a window starting at the left boundary of $-k$ and right boundary of $(t_w-1)s+k+1$. Since the offset between two neighboring subtensors is $st_w$,
we can define the GrateTile configuration as follows

\begin{equation}
\begin{aligned}
\mathcal{G} &= \left\lbrace -k, (t_w-1)s+k+1 \right\rbrace && \pmod {st_w}\\
            &= \left\lbrace -k, k-s+1 \right\rbrace        && \pmod {st_w}
\end{aligned}\,.
\end{equation}

For dilated CNN shown in~(\figref{gratetile_general:b}), a similar process yields $\mathcal{G} = \left\lbrace -kd, kd-s+1 \right\rbrace \pmod {st_w}$.

An interesting property for GrateTile is that any configuration for $\bmod~N$ is also valid for $\bmod~N'$ if $N'|N$.  For example, consider an AlexNet CONV1 whose $(k, s, t_w) = (5, 4, 8)$,
its GrateTile configuration is $\mathcal{G} = \left\lbrace 27, 2 \right\rbrace \pmod {32}$, but $\mathcal{G} = \left\lbrace 3, 2 \right\rbrace \pmod {8}$ is also a valid GrateTile configuration.  In the extreme case, the GrateTile degenerates to \figref{cover0:c} when $N'=1$.  It is thus possible to use a single $N$ across all CNN layers to keep the hardware implementation simple, and we show that $N=8$ can be a suitable choice for most purposes in \secref{eva}.

\subsection{Memory Layout for Compressed Subtensors}\label{sec:propose:compress}
Given a GrateTile configuration, we need to store these subtensors in a data structure that complies with the memory alignment requirement to maximize the benefits of compression.  Since subtensors can have different compressed sizes, we have to store the extra metadata (\eg pointers in \figref{cover1}) separately from the compressed subtensors.  Such metadata are usually too large to fit into the SRAM.
For example, the size of metadata would be $72$~kB for AlexNet CONV2 if each subtensor contains 8 words and requires a 32-bit pointer.
Therefore, a more reasonable choice for metadata storage would be in the DRAM.  However, we must be careful since fetching them would cause extra bandwidth and access latency.

\begin{figure}[t]%
\centering
\subfloat[Uniform division]{\label{fig:grate_compress_large:a}\includegraphics[scale=0.47,page=1]{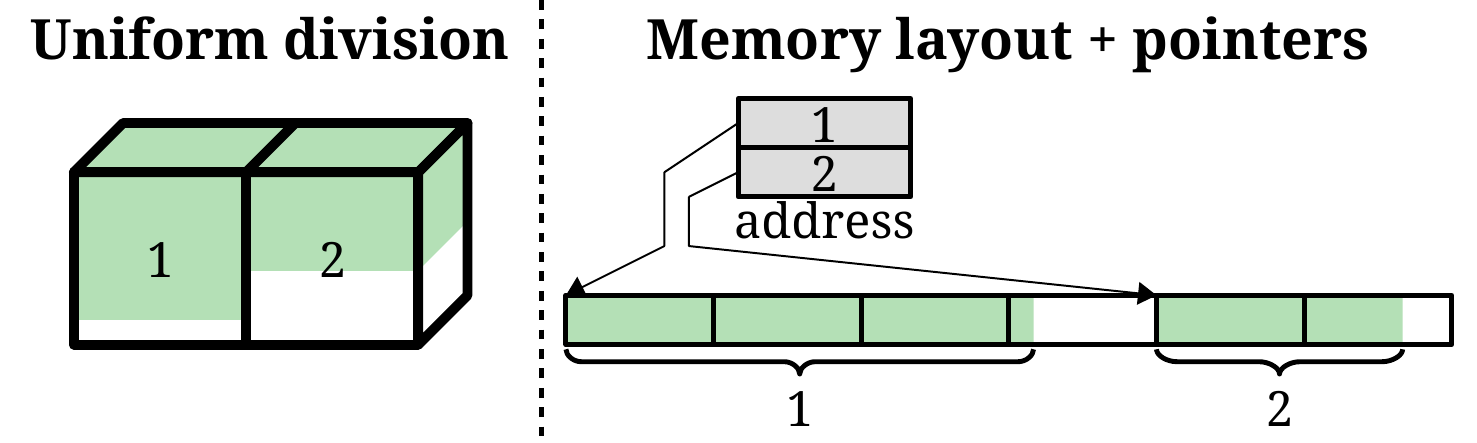}}\\
\subfloat[GrateTile division]{\label{fig:grate_compress_large:b}\includegraphics[scale=0.47,page=2]{propose/grate_compress_large.pdf}}%
\caption{\textbf{The GrateTile data structure. } (a) With uniform division, subtensors align with cache lines, with pointers used to locate the starting addresses. (b) The GrateTile is a near-uniform division; therefore, we extend the uniform division structure by adding the size information for the smaller neighboring subtensors.}\label{fig:grate_compress_large}%
\end{figure}

\figref{grate_compress_large:b} shows how we store the subtensors and metadata.  Since GrateTile is a near-uniform subtensor division methodology, it is relatively straightforward to extend the data structure from uniform division~(\figref{grate_compress_large:a}) for our purposes.
For example, a GrateTile configuration $\mathcal{G} = \left\lbrace 1, 7 \right\rbrace \pmod 8$ is equal to dividing every $8 \times 8$ subtensors further into four small subtensors, and therefore its metadata would extend from the uniform division structure for size 8.  With uniform division, every subtensor has a pointer to the starting address of the subtensor. We extend this structure by adding the compressed sizes of the four smaller neighboring subtensors.  Thus, accessing these subtensors is a two-step procedure, where we first locate the starting address from the pointer and then add the subtensor sizes to get the actual offset for each subtensor.

We now calculate the size of the metadata as follows. As shown in \figref{grate_compress_large:a}, in a uniform subtensor division, we need a pointer for each $8\times8\times8=512$ words.  Since GrateTile only stores these subtensors in aligned addresses, given a $32$-bit addressing space with a 16-byte cache alignment, the size of the pointer is $32-\log_2 16 = 28$.  We now extend this to the GrateTile division and represent the sizes of the four neighboring subtensor by the number of 16-byte caches lines it used.  For this purpose, different GrateTile configurations yield different subtensor sizes, which may require different numbers of bits.  To this end, we select the maximum number among the GrateTile configurations supporting popular CNN kernel sizes.  For kernel sizes 3, 7 and 11, we have $\mathcal{G} = \left\lbrace 1, 7 \right\rbrace \pmod 8$, and a 512-word uniform subtensor divides into four subtensors of sizes 64, 192, 192, and 576 bytes, requiring $3+4+4+6 = 17$ bits of metadata.  For kernel sizes 5 and 9, we have $\mathcal{G} = \left\lbrace 2, 6 \right\rbrace \pmod 8$, which requires $5+5+5+5 = 20$ bits of metadata.  Therefore, for every 512 words of feature map stored in GrateTile format, we need $28+20=48$ bits of metadata, which represents only $0.6$\% of overhead.

\section{Evaluations}\label{sec:eva}

In this section, we discuss the bandwidth reduction with GrateTile in sparse CNN processing compared with several uniform division methods used by other CNN accelerators~\cite{armmlp,cambriconx}.
We simulate memory fetch patterns of representative layers from popular CNN networks~\cite{alexnet,vgg,resnet,vdsr}:

\begin{itemize}
\item \textbf{AlexNet}: All layers, except for the first input layer since it takes dense input images.
\item \textbf{VGG 16}: The layers right before the pooling layers.
\item \textbf{ResNet 18}: The layers right after the pooling layers.
\item \textbf{ResNet 50}: The downsampling CNN layers and the layers before them.
\item \textbf{VDSR}: Every four layers of VDSR, since it consists of $18$ layers of the same shape.
\end{itemize}

\begin{figure}[t]%
\centering
\includegraphics[width=0.5\textwidth,trim={0 3em 0 0}]{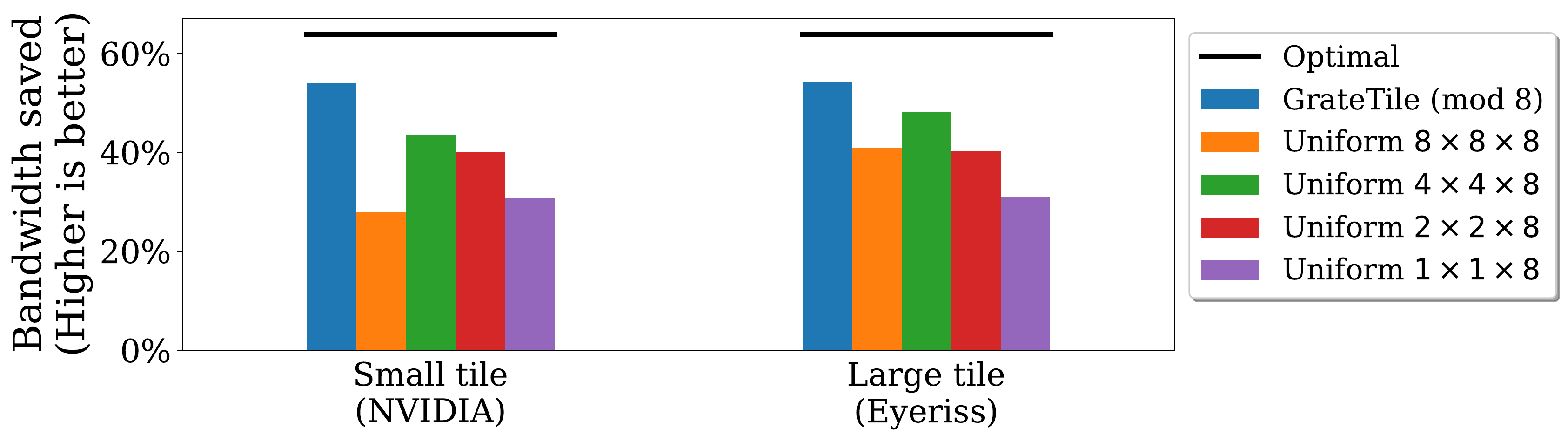}%
\caption{\textbf{Overall bandwidth reduction.} GrateTile provides the best overall bandwidth reduction compared to uniform division on different hardware platforms. Here, the \textit{optimal} bandwidth reduction ratio is defined by the ratio of zero values in the feature map.}\label{fig:saving_geo}
\end{figure}

\figref{saving_geo} illustrates the geometric mean of bandwidth savings from these benchmarks. We use the bitmask compression and the $\bmod~8$ GrateTile configuration for this experiment; we shall discuss the logic behind the selection of this number later.  Note that GrateTile saves an average of 55\% feature map accessing bandwidth, which represents 6-27\% more savings than uniform subtensor division.  We discuss the details of how we arrive at these results in the remainder of this section, as well as insights from our experiments.

\subsection{Experiment Setup}
We perform our simulation on two types of hardware platforms that are characteristics of CNN architectures, namely, an NVIDIA GPU and the Eyeriss architecture.
We assume the memory alignment size is $8$ words ($128$ bits), which is in line with the AXI bus width of~\cite{armmlp}; NVIDIA GPUs also adopt a similar alignment configuration, which is $8$ floating numbers~($256$ bits) per one L1-cache line. To determine the maximum processing tile size, we must consider both double buffering~(prefetching) and convolutional kernels, and assume a reasonable processing tile of less than one-fourth of the buffer size.
Therefore in the following experiments, for an NVIDIA Volta architecture with 64~KB shared memory in one of its processor array, we define the \emph{small tile}~(NVIDIA) configuration to hold a 4K-word feature map subtensor.  For Eyeriss with a 108~KB global buffer, we define the \emph{large tile}~(Eyeriss) configuration to hold 16K words.

\begin{figure*}[t]%
\centering
\subfloat[Bandwidth compression ratio in a small tile platform modeled after NVIDIA Volta.]{\label{fig:saving:b}\includegraphics[width=0.98\textwidth]{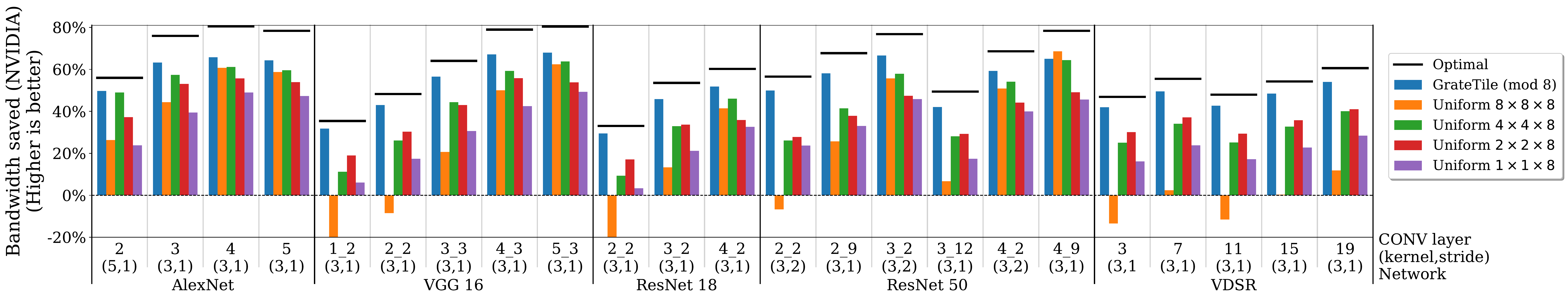}}\\
\subfloat[Bandwidth compression ratio in a large tile platform modeled after Eyeriss.]{\label{fig:saving:c}\includegraphics[width=0.98\textwidth]{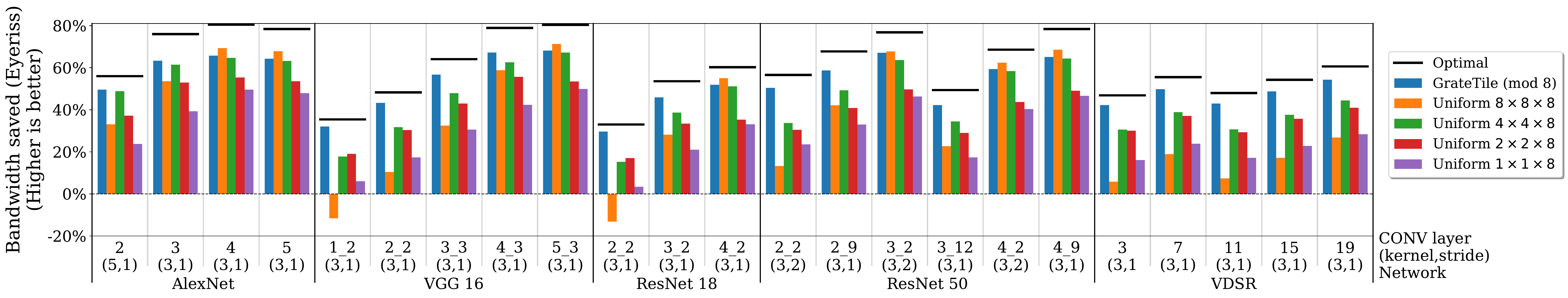}}
\caption{Bandwidth reduction comparison using GrateTile and other subtensor division methods.}\label{fig:saving}
\end{figure*}

\figref{saving} illustrates a bandwidth reduction breakdown of \figref{saving_geo} for individual network layers.  \tabref{tile_cfg} shows the processing tile size and GrateTile configuration for various CNN layers.  We compare GrateTile with uniform subtensor division schemes ranging from $1\times1\times8$ to $8\times8\times8$, under both the small and large tile configurations.  All subtensors are aligned with the cache lines except for the $1\times1\times8$ division, where we compactly the subtensors because each subtensor is too small to fill up one cache line.  \tabref{overhead} and \ref{tab:saving} show the bandwidth overhead caused by fetching the metadata.  

\subsection{Discussions}
From these experiments, we can obtain several useful insights:

\begin{table}[t]%
\centering
\vspace*{-0.3cm}
\caption{GrateTile configurations used in our experiments.}\label{tab:tile_cfg}
\footnotesize
\begin{tabular}{lrrrr}%
\toprule
CNN type & \multicolumn{2}{c}{Tile size modeled after} & GrateTile\\
\cline{2-3}
(kernel,stride) & NVIDIA & Eyeriss & configuration\\
\midrule
$(3,1)$ & $10\stimes18\stimes8$ & $18\stimes18\stimes16$ & $\mathcal{G} = \left\lbrace 1,7 \right\rbrace \pmod 8$\\
$(3,2)$ & $9\stimes17\stimes8$  & $17\stimes17\stimes16$ & $\mathcal{G} = \left\lbrace 0,7 \right\rbrace \pmod 8$\\
$(5,1)$ & $12\stimes20\stimes8$ & $20\stimes20\stimes16$ & $\mathcal{G} = \left\lbrace 2,6 \right\rbrace \pmod 8$\\
\bottomrule
\end{tabular}
\end{table}

\begin{table}[t]%
\centering
\caption{The feature map metadata overhead.}\label{tab:overhead}
\begin{threeparttable}%
\footnotesize
\begin{tabular}{l@{}rr}%
\toprule
Feature map      & \multicolumn{2}{c}{Feature map metadata size}\\
\cline{2-3}
subdivision mode & Bits per KB feature map & Percentage\\
\midrule
GrateTile$\pmod4$    & $(28+20)\times4=192$     & $2.36$\% \\
GrateTile$\pmod8$    & $28+20=48$              & $0.59$\% \\
GrateTile$\pmod{16}$ & $(28+20)\div 4=12$          & $0.15$\% \\
Uniform~$8\stimes8\stimes8$ & $28$                       & $0.34$\% \\
Uniform~$4\stimes4\stimes8$ & $28\times4=112$            & $1.37$\% \\
Uniform~$2\stimes2\stimes8$ & $28\times16=448$           & $5.47$\% \\
Uniform~$1\stimes1\stimes8$ & $32\times64=2048$\tnote{a} & $25.0$\% \\
\bottomrule
\end{tabular}
\begin{tablenotes}
\item [a] The addresses are $32$-bit since we compactly pack each subtensor here.
\end{tablenotes}
\end{threeparttable}
\end{table}

\begin{table}[t]%
\centering
\caption{The impact of metadata on bandwidth reduction.}\label{tab:saving}
\begin{threeparttable}%
\footnotesize
\begin{tabular}{l@{}rrrrr}%
\toprule
Feature map & \multicolumn{5}{c}{Bandwidth saved (\%)}\\
\cline{2-6}
division mode & \multicolumn{2}{c}{Without overhead} && \multicolumn{2}{c}{With overhead}\\
\cline{2-3} \cline{5-6}
& NVIDIA & Eyeriss && NVIDIA & Eyeriss\\
\midrule
GrateTile$\pmod4$           & $46.6$ & $46.6$      && $44.2$ & $44.2$\\
GrateTile$\pmod8$           & $54.7$ & $54.9$      && $\mathbf{54.1}$ & $\mathbf{54.3}$\\
GrateTile$\pmod{16}$        & $56.2$ & ---\tnote{a}&& $\mathbf{56.0}$ & ---\tnote{a} \\
Uniform~$8\stimes8\stimes8$ & $28.4$ & $41.2$      && $27.9$ & $40.9$\\
Uniform~$4\stimes4\stimes8$ & $45.0$ & $49.5$      && $43.6$ & $48.1$\\
Uniform~$2\stimes2\stimes8$ & $45.6$ & $45.8$      && $40.1$ & $40.2$\\
Uniform~$1\stimes1\stimes8$ & $56.5$ & $56.7$      && $30.7$ & $30.9$\\
\bottomrule
\end{tabular}
\begin{tablenotes}
\item [a] In the GrateTile $\pmod{16}$ subtensor division with the small tile configuration~(NVIDIA), a fetched tile is smaller than a subtensor, so GrateTile is not applicable to this case.
\end{tablenotes}
\end{threeparttable}
\end{table}

\textbf{(1) Tile size and bandwidth reduction. }
For uniform tensor division, an optimal division size does not exist because it is a trade-off between the partial tensor accesses and the data indexing overhead.  For example, the larger uniform $8\times8\times8$ division can derive the most benefits by going with larger processing tiles, resulting in a bandwidth improvement of 13\% (40.9\% - 27.9\%).  In comparison, a smaller uniform division like $2\times2\times8$ does not derive similar benefit with larger processing tile (40.2\%-40.1\% = 0.1\%); it also consumes much more metadata than the $8\times8\times8$ division.
Since GrateTile uses a small number of subtensors to prevent partial tensor accesses, it outperforms the best uniform division methods~($4\times4\times8$) according to \figref{saving_geo} and \tabref{saving}.

\textbf{(2) Metadata overhead. }
In \tabref{overhead}, we calculate the metadata required for every $8\times 8\times 8=512$-word feature map and extrapolate the results to different division methods.
In \tabref{saving}, we show the results with and without the bandwidth overhead caused by accessing the metadata.
Observe that without the overhead, the bandwidth saving generally becomes better as the uniform subtensor sizes get smaller.  The only exception is $2\times2\times8$ for large tile configuration due to its cache fragmentation~(\figref{cover1:c}).  The compacted $1\times1\times8$ division can be considered as a performance upper-bound since there is neither partial cache or partial cache accesses, and GrateTile $\bmod~8$ is only $1.8$\% worse than this upper-bound.
However, the $1\times1\times8$ division adds $24.4$\% metadata fetching overhead, making it performs the worst compared with other division methods.

\textbf{(3) Limitations and GrateTile configuration. }
Because GrateTile is best suited for tile-based CNN processing, adopting GrateTile may need to bandwidth overhead by creating unnecessary subtensor division, for example, if an accelerator processes a whole channel before the next channel.  In this scenario, a tile and a feature map have the same sizes at the spatial dimensions, which happens in layers like AlexNet~CONV5 or VGG~16~CONV5\_3 where a uniform division subtensor ($16\times16$) can contain the whole input feature map ($14\times14$).  For these layers, using GrateTile requires 4\% more bandwidth than not dividing the subtensor at all.  It also explains why the $\bmod~16$ GrateTile has slightly better performance (56.0\%-54.1\% = 1.9\%) than $\bmod~8$ in \tabref{saving}.  However, this subtensor division does not work in the smaller tile hardware configuration, which implies a large workspace requirement to compress the subtensors.
Therefore, we claim that the $\bmod~8$ GrateTile is a reasonable choice for most network layers and hardware configurations.

\section{Conclusions}
We propose GrateTile, a hardware-friendly methodology for storing and accessing compressed, sparse feature maps.  GrateTile divides feature maps into uneven subtensors, and in the process, avoids wasteful fetches of partial subtensors and partial cache lines.  Furthermore, it only requires a small metadata indexing overhead to keep track of the locations of the compressed subtensors.  It can be a simple-yet-effective modification for existing CNN accelerators since it is mostly independent of the compression algorithms and requires changes only to the existing feature map division methods.  Our experiments show that GrateTile can save up to $55$\% more bandwidth than the baseline and $6$-$27$\% compared with uniform subtensor division methods.

For hardware compression and decompression, our preliminary SystemVerilog implementation shows promising area efficiency compared to ZRLC, bitmask, and dictionary-based algorithms, with better scalability and less serialization.  We will continue to investigate in this front and share our findings with the community.

\bibliographystyle{IEEEtran}
\bibliography{bare_jrnl}

\end{document}